\UseRawInputEncoding

\documentclass[journal]{IEEEtran}

\ifCLASSOPTIONcompsoc
  \usepackage[nocompress]{cite}
\else
  \usepackage{cite}
\fi

\ifCLASSINFOpdf
\else
\fi

\usepackage{graphicx}
\usepackage{epstopdf}
\usepackage{threeparttable}
\usepackage{multirow}
\usepackage{array}
\usepackage[tight]{subfigure}
\usepackage{caption}
\usepackage{subfig}
\usepackage{setspace}
\usepackage{amsmath}
\usepackage{amsthm}
\usepackage{amssymb}
\usepackage{bm}
\usepackage{etoolbox}
\usepackage{algorithm}
\usepackage{comment}
\usepackage[noend]{algpseudocode}

\usepackage[normalem]{ulem} 
\useunder{\uline}{\ul}{} 

\newcolumntype{P}[1]{>{\centering\arraybackslash}p{#1}}
\newcolumntype{M}[1]{>{\centering\arraybackslash}m{#1}}

\begin{document}
	
\newtheorem{Lemma}{Lemma}
\newtheorem{Definition}{Definition}
\newtheorem{Theorem}{Theorem}

\title{Preventing Dimensional Collapse of Incomplete Multi-View Clustering via Direct Contrastive Learning}

\author{Kaiwu~Zhang,
	Shiqiang~Du,
Baokai~Liu,
and Shengxia~Gao
}


\markboth{IEEE TRANSACTIONS ON XXX}%
{Shell \MakeLowercase{\textit{et al.}}: Bare Demo of IEEEtran.cls for IEEE Journals}

\maketitle

\begin{abstract}
Incomplete multi-view clustering (IMVC) is an unsupervised approach, among which IMVC via contrastive learning has received attention due to its excellent performance. The previous methods have the following problems: 1) Over-reliance on additional projection heads when solving the dimensional collapse problem in which latent features are only valid in lower-dimensional subspaces during clustering. However, many parameters in the projection heads are unnecessary. 2) The recovered view contain inconsistent private information and useless private information will mislead the learning of common semantics due to consistent learning and reconstruction learning on the same feature. To address the above issues, we propose a novel incomplete multi-view contrastive clustering framework. This framework directly optimizes the latent feature subspace, utilizes the learned feature vectors and their sub-vectors for reconstruction learning and consistency learning, thereby effectively avoiding dimensional collapse without relying on projection heads. Since reconstruction loss and contrastive loss are performed on different features, the adverse effect of useless private information is reduced. For the incomplete data, the missing information is recovered by the cross-view prediction mechanism and the inconsistent information from different views is discarded by the minimum conditional entropy to further avoid the influence of private information. Extensive experimental results of the method on 5 public datasets show that the method achieves state-of-the-art clustering results.

\end{abstract}


\IEEEpeerreviewmaketitle

\section{Introduction}\label{sec:introduction}

\IEEEPARstart{I}{n} the current era of information explosion, a large amount of data floods into our lives. How to mine the hidden structure of the data from the massive large-scale data, so that the useful information of the data can be better utilized has gradually attracted widespread attention. As one of the most important unsupervised methods, clustering ~\cite{xu2015unsupervised,Tang2021One,Du2022Enhanc ,zhao2017multi,Du2021Tensor} has long been a key technique in pattern recognition and machine learning, which groups data samples according to a certain criterion so that the similar samples are accumulated into the same cluster. With the development of modern communication technologies, most of the available data are collected from multiple sources or described by various feature collectors, thus generating multi-view data. Since multi-view data can provide common semantics to improve learning efficiency, multi-view clustering~\cite{kumar2011co,nie2016parameter,Xia2022Mul} has received increasing attention.

However, in practical applications, the information of multi-view data may be missing in some views due to the influence of machine damage or sensor problems, which leads to the generation of incomplete multi-view data. The appearance of incomplete multi-view data makes the existing multi-view clustering methods extremely limited and inapplicable. Therefore, in order to deal with the problem of incomplete multi-view data, people have proposed incomplete multi-view clustering methods~\cite{Yin2015Incomplete ,Zhao2016Inc,Li2014Part,Liu2021Ano,Wen2020Inc}. The existing incomplete multi-view clustering methods can be roughly divided into for traditional and deep methods.

Traditional incomplete multi-view clustering methods~\cite{Shao2013Clus,Rai2010Multi,Shao2016Online,Hu2019One} usually utilize traditional machine learning methods to obtain a discriminative new feature representation to replace the original data for clustering. Existing traditional incomplete multi-view clustering methods focus on using zero or mean imputation techniques to fill in incomplete information. In this way, Zhou et al. \cite{Zhou2019Consensus} proposed to fill missing samples with mean eigenvalues, and then perform clustering on the filled data. However, this padding method may introduce some useless or even noisy information, resulting in poor quality of the construction graph. In order to avoid the above problems, people propose to directly learn the common representation or subspace without the process of filling. Hu et al. \cite{Hu2019Dou} proposed a doubly aligned incomplete multi-view clustering (DAIMC) method, which uses regression constraints to align the basis matrices, so that more usable information between views can be obtained. Although traditional incomplete multi-view clustering methods have been well developed, they suffer from the disadvantage of limited representational ability, which leads to their limited performance in complex scenes with real data.

In recent years, deep incomplete multi-view clustering methods~\cite{Jiang2019Dm2c,Wang2018Part,Wen2020Dimc} have gradually become a popular trend in the community due to their strong generalization performance and scalability. Deep incomplete multi-view clustering methods typically utilize an imputation strategy to infer possible values for missing data prior to multi-view clustering. For instance, Xu et al. \cite{Xu2019Adv} proposed using generative adversarial networks to generate desired views for missing views and implemented a weighted adaptive fusion scheme, which can better utilize the complementary information between different views. Due to the rise of contrastive learning, many methods based on contrastive learning have been proposed.  As a typical contrastive learning-based method, COMPLETER \cite{Lin2021cvpr} believed that cross-view consistency learning and data recovery were intrinsically linked within the framework of information theory, and proposed to recover missing data through dual prediction. Both of the above methods learned the consistency information of views by utilizing latent features and applied reconstruction learning and consistency learning to the same features. Since there are both common information of all views and private information for individual view in multi-view data, reconstruction learning hopes to preserve the private information of views as much as possible and consistent learning is dedicated to mining the common information of views. Therefore, these two methods cannot avoid the influence of useless private information in consistency learning. DSIMVC proposed by Tang et al. \cite{Tang2022Deep} effectively solved the above problems by using MLP to learn high-level features after learning latent representations and applying reconstruction loss and contrastive loss on these two features, respectively. In order to prevent the problem that some dimensions of the obtained latent representation do not contain usable information and the model collapses along some dimensions, DSIMVC and other methods~\cite{Xu2022Mul,Li2021CC,Chen2020Asi} via contrastive learning adopt the way of adding projection heads to solve it. But many parameters in the added projection heads are unnecessary, which will increase the complexity of the model.
\begin{figure*}[t]
	\centering
	\includegraphics[width=17cm ]{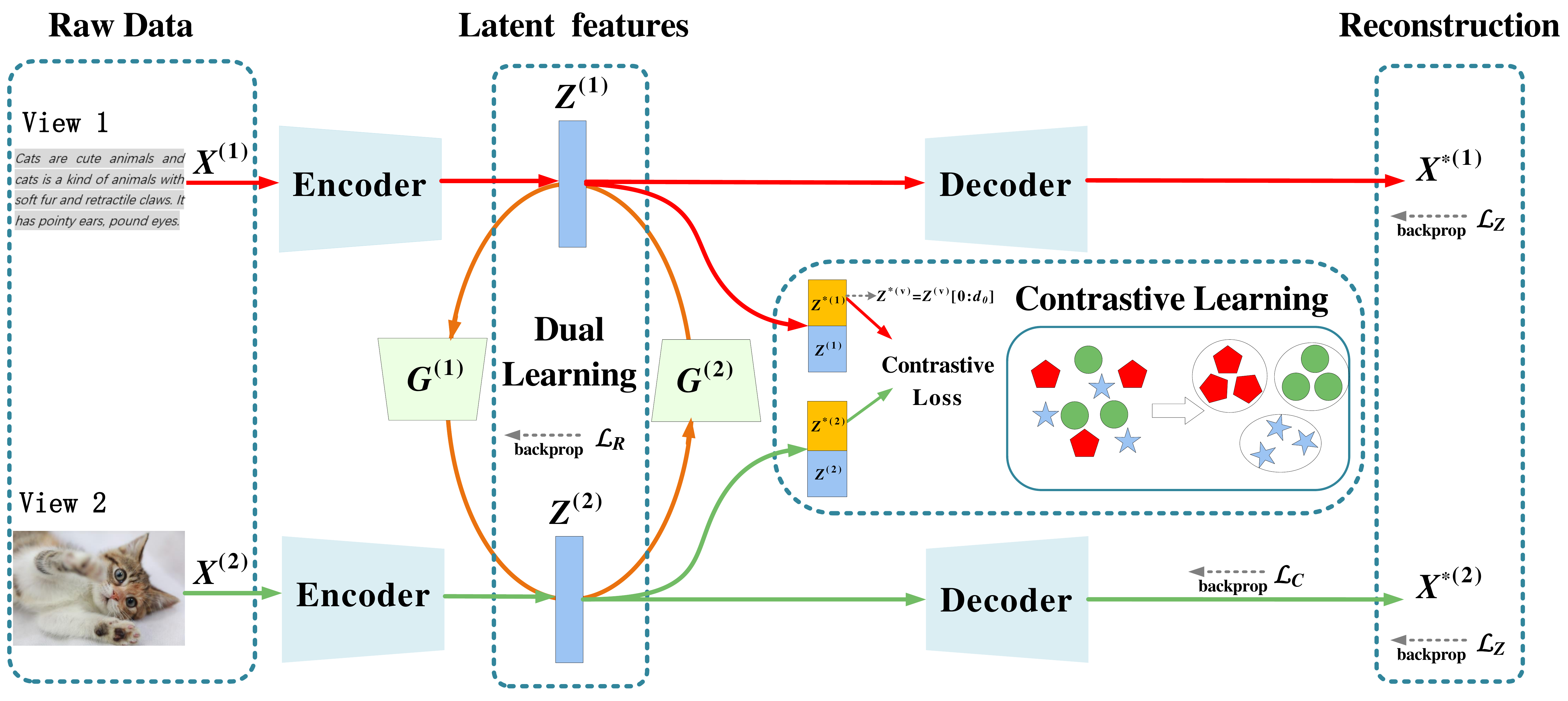}
	\hspace{ 0.1cm}
	\caption{The framework of our model. The features $\mathbf{Z}^v$ are first learned from the raw features~$\mathbf{X}^v$ of each view through the autoencoder. In order to prevent dimensional collapse of latent features when applying contrastive learning, we directly send sub-vectors of latent features to the contrastive loss for learning. The reconstruction objective $\mathcal{L}_{\mathbf{Z}}$ and the consistency objective $\mathcal{L}_{\mathbf{C}}$ are performed on the $\mathbf{Z}^v$ and their sub-vectors, respectively. The $\mathcal{L}_{\mathrm{R}}$ loss aims to recover one view from another through parametric models $G^{(1)}$ and $G^{(2)}$.}
	\label{fig:frame}
\end{figure*}

In summary, although existing incomplete multi-view clustering methods have made important progress by filling missing data with imputation strategies, they suffer from two problems. On the one hand, in order to prevent the phenomenon of dimensional collapse in incomplete multi-view clustering methods based on contrastive learning, existing methods all choose to add projection heads after obtaining latent representations to solve the above problems. However, this method will introduce many unnecessary parameters and reduce the efficiency of the model. Therefore, how to effectively prevent dimensional collapse without increasing the amount of model parameters is the main challenge of incomplete multi-view clustering. On the other hand, some existing incomplete multi-view clustering methods exploit the coherence objective of latent features to explore common semantics across all views. However, these methods usually recover view-private information that is inconsistent across different views when performing data recovery, and further preserve private information in features when the reconstruction objective is required to avoid trivial solutions. As a consequence, how to effectively utilize the consistent information of incomplete data and avoid the influence of view-private information is an crucial challenge for incomplete multi-view clustering.

This paper proposes a novel method for incomplete contrastive multi-view clustering to solve the above problems. Specifically, our main contents include: (1) Design a simple and direct incomplete multi-view clustering model to better utilize the information of latent features of each dimension. Specifically, the autoencoder is first used to learn the latent features of each view from the raw features, and then the sub-vectors of the learned feature vectors are sent to the contrastive loss function, which directly optimizes the representation space and effectively prevents dimensional collapse in the representation space. (2) In order to prevent the performance degradation caused by recovering the inconsistent information of the views, the model adopts the minimization of conditional entropy to recover the missing views, and the inconsistent information will be discarded skillfully. Since reconstruction and consistency learning directly on latent features can cause the above conflicts, unlike most existing methods that require MLP, we adopt a simpler approach where reconstruction learning is implemented on latent features and consistency learning is implemented on their sub-vectors. The approach can achieve better utilization of the useful information of the view at the same level of feature representation. The main contributions of this paper are as follows:

  \begin{itemize}
\item To the best of our knowledge, this is the first work that considers avoiding the effects of dimensional collapse in a way that does not rely on projection heads when applying contrastive learning to clustering. And we design a novel model that optimizes directly in the latent feature subspace, which effectively avoids dimensional collapse in clustering without increasing the amount of unnecessary parameters.

\item We propose an elegant framework for incomplete multi-view contrastive learning, which performs reconstruction learning and consistency learning on the latent features and their sub-vectors respectively. In this way, the conflict between the inconsistency of view-private information and the consistency of common semantics is flexibly resolved. Furthermore, the predictive mechanism of latent feature representation is used to achieve data recoverability.

\item Due to the exquisite design of our model, it has certain robustness in the setting of hyperparameters. Extensive experiments demonstrate that compared with other existing methods, our method has achieved the most advanced effectiveness on 5 real datasets.

\end{itemize}

The rest of this paper is structured as follows. Section \ref{sec:notandrel} introduces some work that is relevant to the content of this paper. Section \ref{sec:pre} provides the main framework and related components of the framework. Section \ref{sec:experiments} describes and analyzes a large number of numerical experiments on real data sets. Finally, we summarize this work in section \ref{sec:conclusion}.

\section{Related Work}\label{sec:notandrel}
In this section, we briefly present recent developments on topics relevant to our work, including two parts: incomplete multi-view clustering and contrastive learning.

\subsection{Incomplete Multi-view Clustering}
As mentioned above, existing incomplete multi-view clustering methods can be divided into traditional incomplete multi-view clustering and deep incomplete multi-view clustering.

Traditional incomplete multi-view clustering methods can be divided into matrix factorization-based methods~\cite{Wen2018Incomp,Shao2015Multi}, kernel learning-based methods~\cite{Liu2018Late,Liu2020Effi,Liu2019Mult}, and graph-based methods~\cite{Wang2019Spec,Li2021Incom,Liu2021Sel}. Matrix factorization-based methods attempt to project incomplete data into a low-dimensionalal common subspace satisfying low-rank constraints by generating latent representations. In this category, partial multi-view clustering (PVC)~\cite{Li2014Part} is one of the earliest works that used non-negative matrix factorization (NMF)~\cite{Lee1999Lear} to obtain a consistent latent representation of incomplete data. Furthermore, MIC~\cite{Shao2015Multi} used the average of the observed samples in the corresponding views to complete unobserved samples, and then performs joint weighted NMF to learn the shared subspace between views. Kernel learning-based methods attempt to recover kernel matrices that are consistent across multiple views in various ways. Liu et al.~\cite{Liu2017Opt} proposed two multi-kernel k-means methods with incomplete kernels. They perform joint imputation on incomplete kernel matrices and combine them for clustering, which is a successful solution with good results. Graph-based methods aim to explore the original geometric structure of the data by exploiting different graphs of all sample relationships. Based on this idea, Wen et al.~\cite{Wen2020Inc} proposed to construct graphs via subspace learning, which fused the partitions generated by each view into a unified matrix.

As far as we know, deep learning has attracted much attention due to its strong scalability and excellent ability to handle large-scale data, and many incomplete multi-view clustering methods have been proposed through various different deep learning frameworks. Based on generative adversarial networks, Wang et al.~\cite{Wang2021Gene} developed a view-specific generative adversarial network with multi-view cycle consistency to generate missing data for one view based on shared representations. Based on the contrastive learning, Yang et al.~\cite{Yang2022Robus} use a noise-robust contrastive loss, which mitigates the effect of false negatives and overcomes the incomplete information challenge.

\subsection{Contrastive Learning}
Contrastive learning~\cite{Chen2020Asi,Wang2020Unde,Liang2022Sema} is a discriminative representation learning framework based on the idea of contrast, which is mainly used for unsupervised representation learning. The main purpose of contrastive learning is to maximize the similarity between instances of the same class and minimize the similarity between instances of different classes. Compared with generative learning, contrastive learning does not need to pay attention to the tedious details of instances, but only needs to learn to distinguish data in the feature space at the abstract semantic level. As such, the generalization ability of the model based on contrastive learning is stronger. Due to its advantages of simple model and saving hardware resources, it has been widely used in the fields of image generation~\cite{Deng2020Dis}, image classification~\cite{He2020MOCO} and clustering~\cite{Xu2022Mul}.

Contrastive learning has currently achieved promising results in computer vision. In terms of image generation, Liu et al.~\cite{Liu2021DivCo} employed contrastive learning to encourage diverse image synthesisers. The proposed DivCo improves the performance of diverse image synthesis without sacrificing the visual quality of the generated images. In terms of representation learning, Kaveh et al. \cite{Kaveh2020Contr} developed a multi-view representation learning method to solve the problem of graph classification through contrastive learning. In terms of clustering, Li et al.~\cite{Li2021CC} proposed a one-stage online clustering method called contrastive clustering, which explicitly performs instance-level and cluster-level contrastive learning. Contrastive learning has become the main research direction of unsupervised learning because of its excellent achievements in various fields.

\section{Proposed Method}
\label{sec:pre}
In this section, we introduce each implementation detail of the proposed incomplete contrastive multi-view clustering model, including the within-view reconstruction module, multi-view contrastive learning module, and missing-view data recovery module.

\subsection{Within-view Reconstruction Module}
We treat the dataset as the original feature, where the dataset has multiple views with $K$ common clusters. For multi-view data, the existing methods generally try to learn the salient features of the sample from the information contained in the original data. In particular, deep autoencoders can accomplish the above tasks and have been applied in many research fields due to their excellent performance. In this paper, we use $\left\{\mathbf{X}^v \in \mathbb{R}^{N \times M_v}\right\}_{v=1}^ V$ to represent a multi-view dataset with $V$ views and $N$ samples, and define $\mathbf{x}_i^v \in \mathbb{R}^{M_v}$ to represent the $M_v$-dimensional samples from the $v$-th view. In order to obtain the feature information of each view, we use autoencoder to transform heterogeneous multi-view data into clustering-friendly features. Specifically, the feature representation $\left\{\mathbf{Z}^v\right\}_{v=1}^V$ of $v$-th view is learned by encoder $f_{ec}^v\left(\mathbf{X}^v ; \theta^v\right)$ and decoder $f_{dc}^v\left(\mathbf{Z}^v ; \phi^v\right)$, where $\theta^v$ and $\phi^v$ are network parameters, and $\mathbf{z}_i^v=f_{ec}^v\left(\mathbf{x}_i^v\right) \in \mathbb{R}^D$ as the $D$-dimensionalal latent feature of the $i$-th sample. Therefore, in the $v$-th view, the reconstruction loss between the encoder input $\mathbf{X}^v$ and the decoder output ${\mathbf{X}^*}^v \in \mathbb{R}^{N \times M_v}$ can be defined as $\mathcal{L}_{\mathrm{Z}}^v$, so the reconstruction loss of all views can be expressed as:
\begin{align}
\label{eq:x1}
\mathcal{L}_{\mathrm{Z}}=\sum_{v=1}^V \mathcal{L}_{\mathrm{Z}}^v=\sum_{v=1}^V \sum_{i=1}^N\left\|\mathbf{x}_i^v-f_{dc}^v\left(f_{ec}^v\left(\mathbf{x}_i^v\right)\right)\right\|_2^2
\end{align}

The representation $\mathbf{Z}^v$ is defined by $\mathbf{Z}^v=f_{ec}^v\left(\mathbf{X}^v\right)$. The deep multi-view clustering methods based on autoencoders generally aimed to learn cluster-friendly common representations by maximizing the consistency between latent features of different views. However, the existing multi-view clustering methods have the following two issues: (1) Most methods suffer from the dimensional collapse problem that the latent feature vector learned in the model collapses along some dimensions so that the dimension of the embedding vector cannot be fully utilized. Although some methods based on contrastive learning solve the above problems by adding additional projection heads, this approach will result in increased computation and many parameters in the projection heads are unnecessary. (2) Some multi-view clustering methods perform reconstruction learning and consistency learning on the same latent features. The purpose of reconstruction learning is to enable the learned latent features to contain more view-private information of the original features, while consistency learning hopes to mine useful common information between views, which will lead to poor quality of the learned latent features and affect the final clustering results.

To avoid the above problems, we design a multi-view contrastive learning module. Inspired by~\cite{Jing2022Unde}, the projector in contrastive learning is of great significance to prevent dimensional collapse, and a linear projector only need to satisfy the diagonal and low rank to effectively prevent dimensional collapse. The details can be found in \cite{Jing2022Unde}. Accordingly, we adopt a novel way to feed the sub-vectors of the latent representation $\mathbf{Z}^v$ directly into the spectral contrastive loss, which satisfies the above conditions and optimizes the representation space, thus preventing the dimensional collapse that occurs in the clustering model. At the same time, since we perform consistent learning on carefully selected sub-vectors, the remaining irrelevant information does not participate in the contrastive learning, which helps to filter out the view-private information of features and obtain higher-quality common representations. The specific content of this part is elaborated in the next subsection.
\subsection{Multi-view Contrastive Learning Module}
It is obvious that the latent feature $\left\{\mathbf{Z}^v\right\}_{v=1}^V $ of each view obtained from the original features does not separate the consistent information and private information of each view. Most existing methods directly perform contrastive learning on features $\left\{\mathbf{Z}^v\right\}_{v=1}^V$, which causes private information in features to interfere with consistent learning. To solve this problem, Xu et al.~\cite{Xu2022Mul} avoid the above situation by adding MLP after the feature $\left\{\mathbf{Z}^v\right\}_{v=1}^V$ to obtain advanced features. Although this method can resolve the conflict between the reconstruction objective and the consistency objective, the addition of MLP increases the complexity of the model and reduces the efficiency of the model. Therefore, we adopt a simple way to use sub-vectors representing features and feed the sub-vectors directly into the contrastive loss to solve the above problem, where we utilize the $\left[0:d_0\right]$-dimensional vectors of latent features as sub-vectors for consistency learning.

Since the purpose of contrastive learning is to maximize the similarity between positive sample pairs and minimize the similarity between negative sample pairs, the sub-vector $\mathbf{z}_i^{*v}$ defined for each feature in our model has $(V N - 1)$ feature pairs, where $\left\{\mathbf{z}_i^{*v}, \mathbf{z}_i^{*n}\right\}_{n \neq v}$ are $(V-1)$ positive pairs and the rest $V(N-1)$ pairs are negative pairs. Based on the spectral contrastive loss \cite{Hao2021Prova}, we adopt the following loss to achieve the consistency objective, so that the features focus on learning the common semantics of all views:
\begin{align}
\label{eq:x2}
\mathcal{L}_{\mathrm{C}}=& \sum_{v=1}^V \sum_{n=v+1}^N\left[-\frac{2}{N} \sum_{i=1}^{N} {\mathbf{z}_i^{*v}}^{\top} {\mathbf{z}_i^{*n}}\right.\nonumber\\
&\left.+\frac{1}{2 C_N^2} \sum_{i=1}^{N} \sum_{j \neq i}\left({\mathbf{z}_i^{*v}}^{\top}{\mathbf{z}_j^{*v}}\right)^2\right]
\end{align}
where $C_N^2$ denotes the combinatorial operations.

The feature representation $\mathbf{Z}^v$ obtained by our designed network avoids model collapse through direct contrast learning on its sub-vector and retains more useful information by further optimizing the representation space. In the case of complete data, we can get a cluster-friendly representation through the above steps and run k-means on the feature representation to get good results. Whereas, most of the data in practical applications are incomplete. Such incomplete data will greatly affect the quality of the learned latent representation and make multi-view clustering methods limited. Accordingly, researchers have proposed incomplete multi-view clustering methods to deal with incomplete data. Most of the current incomplete multi-view clustering methods use the mean filling method to solve the problem of missing information. But this method introduces noise information, and the learned representations cannot fully utilize the consistent information of views. To address this issue, we attempt to use a missing view recovery module to predict and recover missing views from the existing available views, so that our method can reduce the impact of missing information on clustering performance.

\subsection{Missing View Data Recovery Module}
In this module, for the latent feature structure of the original data obtained by the autoencoder, we encourage that $\mathbf{Z}^p$ can be determined by $\mathbf{Z}^q$ by minimizing the conditional entropy $H\left(\mathbf{Z}^p \mid \mathbf{Z}^q\right)$. This mechanism enables the missing views to be determined by the available views, which can not only achieve the representation recovery of the missing views but also cleverly discard the inconsistent information between different views to obtain a better consistent representation. We represent the conditional entropy by the following equivalent log conditional likelihood transformation:
\begin{align}
\label{eq:x3}
\begin{aligned}
H\left(\mathbf{Z}^p \mid \mathbf{Z}^q\right)=-\mathbb{E}_{\mathcal{Z}_{\mathbf{z}^p, \mathbf{z}^q}}\left[\log \mathcal{P}\left(\mathbf{Z}^p \mid \mathbf{Z}^q\right)\right]
\end{aligned}
\end{align}

However, it is difficult to directly estimate the expected value in equation (3). A general approach to approximate this objective is by introducing a variational distribution $\mathcal{Q}\left(\mathbf{Z}^p \mid \mathbf{Z}^ q \right)$ and maximize the lower bound of the  expectations, i.e., $\mathbb{E}_{\mathcal{P}_{\mathbf{Z}^p, \mathbf{Z}^q}}\left[\log \mathcal{Q}\left(\mathbf{Z}^p \mid \mathbf{Z}^q\right)\right]$.

The above $\mathcal{Q}$ has no strong restrictions and can be implemented in various ways, such as Gaussian distribution or through neural networks. For simplicity and directness, we use Gaussian distribution and utilize $G(\cdot)$ as a parameter function to map $\mathbf{Z}^p$ to $\mathbf{Z}^q$, i.e., $\mathcal {N} \left(\mathbf{Z}^p \mid G^{(q)}\left(\mathbf{Z}^q\right), \sigma \mathbf{I}\right)$, the $\sigma \mathbf{I}$ appearing here is a diagonal matrix. By ignoring the constants derived from the Gaussian distribution, the objective function in our data recovery can be defined as:
\begin{align}
\label{eq:x4}
\mathcal{L}_{\mathrm{R} } = \max -\mathbb{E}_{\mathcal{P}_{\mathbf{Z}^p, \mathbf{Z}^q}}\left\|G^{(q)}\left(\mathbf{Z}^q\right)-\mathbf{Z}^p\right\|_2^2
\end{align}

In order to visually observe the objective function of the data recovery part, for the data of the two views, the loss function has the following form:
\begin{align}
\label{eq:x5}
\mathcal{L}_{\mathrm{R} }&=\left\|\left(f_{ec}^2\left(\mathbf{X}^2\right)\right)-G^{(1)}\left(f_{ec}^1\left(\mathbf{X}^1\right)\right)\right\|_2^2 \nonumber \\
&+\left\|\left(f_{ec}^1\left(\mathbf{X}^1\right)\right)-G^{(2)}\left(f_{ec}^2\left(\mathbf{X}^2\right)\right)\right\|_2^2
\end{align}

Through the above loss optimization model, the missing view data recovery module can better recover complete information from another view, further reducing the impact of missing data on the model. The missing information can be recovered from such a way, i.e., $\mathbf{Z}^2=G^{(1)}\left(\mathbf{Z}^1\right)=G^{(1)}\left(f_{ec}^1\left(\mathbf{X}^1\right)\right)$ and $\mathbf{Z}^1=G^{(2)}\left(\mathbf{Z}^2\right)=G^{(2)}\left(f_{ec}^2\left(\mathbf{X}^2\right)\right)$.



\begin{table*}[t]
	\caption{Details of widely used benchmark data sets.}
	\label{tab:cluster1}\centering
\renewcommand\tabcolsep{4.2pt}
	\begin{tabular}{cccccc}  \hline
		\cline{1-6}
		Data  &BDGP & MNIST-USPS      & Caltech  & Fashion     & Noisy MNIST              \\ \hline
		view1 & 1750  & 784    &40 & 784   & 784    \\
		view2 & 75  & 784      & 254& 784  & 784       \\
		size   & 2500  & 5000  & 1400   & 10000    & 20000              \\
		classes & 5   & 10    & 7    & 10         & 10                     \\ \hline
	\end{tabular}
\end{table*}

\section{Experiments}
\label{sec:experiments}
In this section, we compare the proposed method with 7 existing state-of-the-art methods on 5 widely used multi-view datasets, while using different missing rates on different datasets. To demonstrate that our method can achieve missing view recovery, a visualization of missing view recovery is added in the experimental section to further illustrate the effectiveness of our method.  All simulations are implemented using PyTorch 1.7.1, and the original code of this method will be provided later.
\subsection{Experimental Setup}
In our experiments, we employ randomly generated incomplete multi-view datasets to validate our algorithm. For each dataset, we generate incomplete samples by randomly removing partial views across all samples. The missing rate $\eta$ is defined as $\eta=(n-m) / n$, where $m$ is the number of complete examples, and $n$ is the number of the whole dataset, which ranges from 0.1 to 0.7 with an interval of 0.2.
\subsubsection{Datasets Descriptions}
\begin{table*}[t]
	\caption{Clustering results ($\%$) on BDGP.}
	\label{tab:cluster2}\centering
\renewcommand\tabcolsep{4.2pt}
	\begin{tabular}{ccccccccccccc}  \hline
		\cline{1-13}
	                &\multicolumn{4}{c}{ACC}                &\multicolumn{4}{c}{NMI}                 &\multicolumn{4}{c}{ARI}               \\ \hline
	Method          &0.1   &0.3 &0.5 &0.7  &0.1   &0.3 &0.5 &0.7   &0.1   &0.3 &0.5 &0.7     \\ \hline
AE$^2$-NET (CVPR'19)         &57.38   &54.03 &49.91 &46.88  &46.84   &42.00 &38.44 &35.21   &18.12   &12.74 &9.86  &7.39  \\
  MFLVC (CVPR'22)            &97.18   &84.33 &78.07 &57.14  &91.84   &73.64 &62.91 &38.25   &93.12   &70.55 &57.04 &26.26  \\
	UEAF (AAAI'19)           &92.44   &89.48 &86.92 &79.92  &79.26   &72.66 &67.56 &55.70   &82.08   &75.59 &70.21 &56.85 \\
	FLSD (TCYB'20)           &90.88   &80.64 &79.08 &75.60  &75.03   &58.56 &56.44 &56.01   &78.60   &59.29 &57.18 &51.80 \\
CDIMC-net (IJCAI'20)          &82.52   &73.20 &61.72 &52.76  &70.15   &66.52 &41.78 &46.13   &62.95   &57.26 &37.15 &29.60  \\
COMPLETER (CVPR'21)          &52.65   &51.10 &43.86 &41.66  &44.00   &42.52 &37.09 &31.76   &23.66   &16.83 &12.25 &7.61  \\
DIMVC (AAAI'22)              &95.90   &94.75 &93.24 &89.01  &88.00   &85.32 &81.37 &73.61   &90.11   &87.39 &83.94 &75.00 \\
  OURS           &\pmb{97.97}   &\pmb{96.86} &\pmb{95.43} &\pmb{91.36}  &\pmb{93.90}   &\pmb{90.49} &\pmb{86.60} &\pmb{78.12}   &\pmb{95.02}   &\pmb{92.33} &\pmb{88.98} &\pmb{79.83}  \\                                                                             \hline
	\end{tabular}
\end{table*}

\begin{table*}[t]
	\caption{Clustering results ($\%$) on MNIST-USPS.}
	\label{tab:cluster3}\centering
\renewcommand\tabcolsep{4.2pt}
	\begin{tabular}{ccccccccccccc}  \hline
		\cline{1-13}
	                &\multicolumn{4}{c}{ACC}                &\multicolumn{4}{c}{NMI}                 &\multicolumn{4}{c}{ARI}               \\ \hline
	Method          &0.1   &0.3 &0.5 &0.7  &0.1   &0.3 &0.5 &0.7   &0.1   &0.3 &0.5 &0.7     \\ \hline
AE$^2$-NET (CVPR'19)         &74.42   &64.63 &50.51 &40.19  &72.53   &61.58 &46.65 &37.47   &63.76   &48.70 &29.35 &19.49 \\
  MFLVC (CVPR'22)            &92.28   &71.28 &61.26 &23.62  &85.72   &60.14 &40.44 &20.74   &82.69   &40.60 &31.24 &3.88  \\
  	UEAF (AAAI'19)           &74.76   &71.20 &69.62 &60.12  &69.58   &64.65 &59.99 &51.73   &61.53   &56.10 &51.72 &41.70 \\
	FLSD (TCYB'20)           &79.16   &70.77 &68.28 &65.96  &75.83   &68.03 &63.64 &60.15   &68.70   &58.60 &54.11 &50.61 \\
CDIMC-net (IJCAI'20)          &67.10   &62.96 &62.60 &52.04  &72.18   &65.09 &68.43 &54.08   &56.28   &53.82 &49.50 &32.29 \\
COMPLETER (CVPR'21)          &88.60   &87.94 &82.96 &84.64  &92.83   &90.94 &87.11 &82.65   &85.81   &85.00 &77.86 &75.99 \\
DIMVC  (AAAI'22)             &85.44   &79.37 &78.64 &78.22  &78.14   &77.90 &77.93 &76.14   &73.57   &70.83 &70.17 &70.33 \\
  OURS           &\pmb{99.40}   &\pmb{98.73} &\pmb{97.48} &\pmb{95.94}  &\pmb{98.23}   &\pmb{96.52} &\pmb{93.54} &\pmb{90.47}   &\pmb{98.66}   &\pmb{97.20} &\pmb{94.49} &\pmb{91.17}  \\                                                                             \hline
	\end{tabular}
\end{table*}

The experiments in this paper selected 5 widely-used data sets for testings, the statistics of which are shown in Table \ref{tab:cluster1}. The specific details of the dataset are as follows:
\begin{enumerate}
	\item   BDGP~\cite{Cai2012Joi} dataset: BDGP is a dataset of images of drosophila embryos. It contains 2,500 samples of 5 classes, where each sample has two views of 1750-dimensional visual features and 79-dimensional textual features.
	\item  MNIST-USPS~\cite{Peng2019COMI} dataset: MNIST-USPS contains 5,000 samples. It has a total of 10 categories and two views, where the first view is obtained from the MNIST dataset and the second view is obtained from the USPS dataset.
	\item Caltech~\cite{Li2004Learni} dataset: Caltech is an image dataset. We construct a dual-view dataset based on its multi-view to evaluate our method and compare algorithms.
	\item  Fashion~\cite{Xiao2017 Fa} dataset:  Fashion is a 10,000-sample image dataset about products, and we treat two different styles in the dataset as two views of a product by using.
	\item Noisy MNIST~\cite{Wang2015Onde} dataset: This dataset is the MNIST dataset with multi-views. It uses randomly selected intra-class images with Gaussian noise and the original MNIST image as its two views. In order to verify that the proposed method can handle large-scale image datasets, this paper uses a Noisy MNIST subset with 20,000 samples for experiments.

\end{enumerate}
\subsubsection{Comparison Algorithms}
We verify the performance of the proposed method by comparing our algorithm with state-of-the-art multi-view clustering methods. The specific information of these seven algorithms is as follows:
\begin{enumerate}
	\item UEAF~\cite{Wen2019Uni}  uses graph learning to mine the local structure of the data, and simultaneously considers the reconstruction of the hidden information of missing views and the adaptive importance assessment of different views in one framework.
	\item FLSD~\cite{Wen2020Gene} introduces semantic consistency constraints and proposes a graph regularized matrix factorization model, which simultaneously considers the latent and shared representations of different views.
	\item AE$^2$-NET~\cite{Zhang2019AE} is to encode the internal information from heterogeneous views into a comprehensive representation, and this model can automatically balance the complementarity and consistency of different views.
	\item MFLVC~\cite{Xu2022Mul} proposes a multi-level feature learning framework for contrastive multi-view clustering. The model can learn different levels of features and realizes reconstruction learning and consistency learning in a fusion-free
manner.
	\item CDIMC-net~\cite{Wen2020CDIMC} captures latent features of views by combining deep encoders and graph embedding strategies into a model framework.
	\item COMPLETER~\cite{Lin2021cvpr} is the first work that provides a theoretical framework to unify representation learning and data recovery.
	\item DIMVC~\cite{Xu2022Dee} proposes a deep incomplete multi-view model without interpolation and fusion. DIMVC provides an implementation of a high-dimensionalal mapping and transforms this complementary information into supervised information with high confidence to achieve multi-view clustering consistency for complete and incomplete data.
\end{enumerate}

\begin{table*}[t]
	\caption{Clustering results ($\%$) on Caltech.}
	\label{tab:cluster4}\centering
\renewcommand\tabcolsep{4.2pt}
	\begin{tabular}{ccccccccccccc}  \hline
		\cline{1-13}
	                &\multicolumn{4}{c}{ACC}                &\multicolumn{4}{c}{NMI}                 &\multicolumn{4}{c}{ARI}               \\ \hline
	Method          &0.1   &0.3 &0.5 &0.7  &0.1   &0.3 &0.5 &0.7   &0.1   &0.3 &0.5 &0.7     \\ \hline
AE$^2$-NET (CVPR'19)         &46.54   &36.98 &18.81 &17.44  &37.51   &23.36 &1.71  &1.09    &28.20   &13.29 &5.3   &1.6 \\
  MFLVC (CVPR'22)            &60.99   &46.27 &38.86 &36.36  &52.06   &35.56 &28.39 &25.31   &43.16   &25.19 &16.70 &12.64 \\
	UEAF (AAAI'19)           &41.01   &35.57 &35.00 &34.14  &26.36   &22.83 &22.03 &21.07   &19.09   &16.70 &11.67 &13.82 \\
	FLSD (TCYB'20)           &46.07   &42.57 &41.74 &40.30  &31.57   &27.37 &26.09 &25.49   &24.88   &20.02 &19.41 &19.24 \\
CDIMC-net (IJCAI'20)          &50.79   &49.07 &44.29 &42.93  &49.29   &42.75 &41.65 &38.48   &35.15   &33.97 &28.55 &26.78 \\
COMPLETER (CVPR'21)          &47.24   &46.73 &47.53 &44.99  &45.93   &44.22 &44.72 &43.12   &27.49   &27.28 &24.59 &24.26 \\
DIMVC  (AAAI'22)             &46.88   &43.96 &37.25 &36.30  &32.90   &26.72 &23.58 &20.03   &25.40   &19.25 &14.73 &12.99 \\
  OURS           &\pmb{67.51}   &\pmb{56.47} &\pmb{55.93} &\pmb{53.66}  &\pmb{56.98}   &\pmb{51.26} &\pmb{47.55} &\pmb{44.15}   &\pmb{47.36}   &\pmb{42.29} &\pmb{39.83} &\pmb{37.54}  \\                                                                             \hline
	\end{tabular}
\end{table*}

\begin{table*}[t]
	\caption{Clustering results ($\%$) on Fashion.}
	\label{tab:cluster5}\centering
\renewcommand\tabcolsep{4.2pt}
	\begin{tabular}{ccccccccccccc}  \hline
		\cline{1-13}
	                &\multicolumn{4}{c}{ACC}                &\multicolumn{4}{c}{NMI}                 &\multicolumn{4}{c}{ARI}               \\ \hline
	Method          &0.1   &0.3 &0.5 &0.7  &0.1   &0.3 &0.5 &0.7   &0.1   &0.3 &0.5 &0.7     \\ \hline
AE$^2$-NET (CVPR'19)         &65.52   &59.37 &50.52 &43.97  &65.09   &59.40 &51.13 &44.04   &52.99   &44.37 &31.71 &23.53\\
  MFLVC (CVPR'22)            &88.75   &80.35 &52.72 &39.93  &79.84   &69.57 &50.59 &36.85   &75.34   &59.29 &26.92 &10.66 \\
  	UEAF (AAAI'19)            &60.70   &51.89 &40.59 &31.07  &62.73   &52.81 &40.86 &29.34   &47.68   &35.80 &22.35 &14.61 \\
	FLSD (TCYB'20)           &66.57   &65.06 &63.66 &56.02  &69.43   &66.87 &64.65 &62.15   &54.81   &52.06 &49.70 &45.41 \\
CDIMC-net (IJCAI'20)          &65.42   &51.05 &46.32 &43.47  &71.67   &69.08 &50.01 &48.72   &55.17   &40.19 &30.81 &29.45 \\
COMPLETER (CVPR'21)          &78.55   &75.99 &72.97 &71.91  &84.76   &80.64 &78.14 &74.05   &71.33   &75.34 &67.32 &62.99 \\
DIMVC (AAAI'22)              &64.49   &62.60 &62.65 &61.43  &75.81   &70.51 &69.70 &64.43   &58.49   &54.93 &53.78 &49.58 \\
  OURS           &\pmb{96.53}   &\pmb{94.14} &\pmb{90.98} &\pmb{86.94}  &\pmb{92.86}   &\pmb{88.82} &\pmb{84.23} &\pmb{79.19}   &\pmb{92.67}   &\pmb{87.93} &\pmb{82.09} &\pmb{74.98}  \\                                                                             \hline
	\end{tabular}
\end{table*}

\subsubsection{Evaluation Metrics}
The clustering performance of all methods was measured by the following widely used evaluation metrics: Accuracy (ACC), Normalized Mutual Information (NMI) and ARI. The larger the value of all indicators, the better the clustering performance. The specific information of these three evaluation indicators is as follows:
\begin{enumerate}
	\item Accuracy (ACC). ACC is used to compare the obtained labels with the true labels provided by the data. ACC is defined as follows:
	\begin{align}\label{eq:rtlrr0}
	& \mathrm{ACC}=\frac{\sum_{i=1}^{n}\delta(t_{i},map(h_{i}))}{n}, &
	\end{align}
where $t_{i}$ represents the true label of the data. $h_{i}$ represents the label of the $i$-th sample obtained by clustering, and $map(h_{i})$ represents the reproduction allocation of the best clustering index to ensure the correctness of statistics. $\delta(\cdot,\cdot)$ is the indication function, which is defined as:

	\begin{align}
	\label{E4}
	&\delta(u,v)=\left\{
	\begin{array}{ll}
	1, \ & u=v,\\
	0, \ & otherwise.
	\end{array}\right.&
	\end{align}
	\item Normalized Mutual Information (NMI). NMI is defined as follows:
\begin{equation}
\operatorname{NMI}=\frac{2 \sum_{i=1}^k \sum_{j=1}^{\hat{k}} \frac{m_{i j}}{m} \log \frac{m_{i j} m}{\sum_{i=1}^k q_i \sum_{j=1}^k p_j}}{-\sum_{i=1}^k \frac{q_i}{m} \log \frac{q_i}{m}-\sum_{j=1}^{\hat{k}} \frac{p_j}{m} \log \frac{p_j}{m}}
\end{equation}
where $m$ is the total number of samples and $k$ is the number of clusters. $q_i$ and $p_j$ are the number of the obtained clustering results and true labels respectively.
	\item ARI. ARI is defined as follows:
	\begin{align}\label{eq:rtlrr0}
	&  \mathrm{ARI}=\frac{\sum_{i, j=1}^c C_{n_{i j}}^2-\frac{\sum_{i=1}^c C_{s_i}^2 \sum_{i=1}^c C_{t_i}^2}{C_n^2}}{\frac{1}{2}\left(\sum_{i=1}^c C_{s_i}^2+\sum_{i=1}^c C_{t_i}^2\right)-\frac{\sum_{i=1}^c C_{s_i}^2 \sum_{i=1}^c C_{t_i}^2}{C_n^2}}.&
	\end{align}
 $C_n^m$ represents the number of $n$ samples selected from $m$ samples.

\end{enumerate}

\subsection{Experimental Results and Analysis}

For all compared algorithms, we use default parameters or iteratively search in a given parameter set to find the best parameters for these methods and report their best clustering results. For the clustering method that can only be applied to the complete multi-view data~(AE$^2$-NET and MFLVC) adopt the same way as~\cite{Lin2021cvpr}, we use the average value of the feature values of the corresponding views to fill in the missing views, and cluster on the completed data set obtained after filling. For each dataset, all methods were performed on the same randomly formed incomplete cases and their best results were reported for fair comparison. The clustering metrics (ACC, NMI and ARI) comparison is presented in Table \ref{tab:cluster2}-\ref{tab:cluster6}. From these tables, we obtain the following observations:

1) From the Table \ref{tab:cluster2}-\ref{tab:cluster6}, it can be observed that the clustering metrics (ACC, NMI and ARI) gradually decrease as the missing rate increases. The performance of other incomplete multi-view clustering methods is better than the complete multi-view clustering method when the missing rate is large, which indicates that the imputed views inferred by other incomplete multi-view clustering methods contain more semantic information than the average vectors and thus alleviate the risk of degraded clustering performance due to  the adverse effects of noise introduced by mean value filling.

2) The clustering results of the proposed method on all data sets are obviously better than other methods, especially when dealing with large-scale data sets and high miss rate. For example, on Noisy MNIST with a missing rate of 0.7, our method exceeds the second best value by approximately 7.5\%, 5.2\% and 7.1\% in terms of ACC, NMI and ARI, respectively. And since our method can effectively prevent the effect of dimensional collapse by using direct contrastive learning, it can more explicitly capture the useful information contained in latent features, thus greatly improving the clustering performance.

3) The clustering performance of all algorithms decreases with increasing missing rate, but the least decrease is clearly observed for our proposed method. Especially on the MNIST-USPS dataset, the performance drop of the comparison algorithm is larger than ours, and the ACC of our method only drops 3.46\% in the case of missing rate 0.1-0.7. These observations proved that our algorithm in the case of a larger missing rate can still normal processing data. This reflects that our recovered views are more conducive to clustering, and our method discards inconsistent information between views so that it can better utilize views to obtain consistent information. This advantage allows our method to perform clustering tasks well even when many view data are missing.

\begin{table*}[t]
	\caption{Clustering results ($\%$) on Noisy MNIST. '-' indicates that the experimental results could not be obtained due to insufficient memory, and the best results of the first place are shown in bold.}
	\label{tab:cluster6}\centering
\renewcommand\tabcolsep{4.2pt}
	\begin{tabular}{ccccccccccccc}  \hline
		\cline{1-13}
	                &\multicolumn{4}{c}{ACC}                &\multicolumn{4}{c}{NMI}                 &\multicolumn{4}{c}{ARI}               \\ \hline
	Method          &0.1   &0.3 &0.5 &0.7  &0.1   &0.3 &0.5 &0.7   &0.1   &0.3 &0.5 &0.7     \\ \hline
AE$^2$-NET (CVPR'19)         &36.25   &28.07 &27.91 &26.07  &30.48   &21.22 &19.93 &16.08   &18.25   &10.50 &7.49  &4.61\\
  MFLVC (CVPR'22)            &89.99   &71.29 &43.08 &23.94  &79.48   &51.04 &38.51 &16.95   &78.79   &45.69 &15.32 &4.15 \\
	UEAF (AAAI'19)            &-   &- &- &-  &-   &- &- &-   &-   &- &- &- \\
	FLSD (TCYB'20)           &-   &- &- &-  &-   &- &- &-   &-   &- &- &- \\
CDIMC-net (IJCAI'20)          &40.09   &38.09 &34.72 &27.86  &48.33   &43.83 &32.85 &22.99   &30.89   &27.22 &21.30 &12.17 \\
COMPLETER (CVPR'21)          &89.99   &87.51 &80.94 &74.41  &86.05   &82.36 &76.52 &67.35   &82.85   &79.17 &71.87 &61.83 \\
DIMVC (AAAI'22)              &64.93   &58.56 &51.57 &48.88  &61.57   &60.46 &48.97 &47.17   &49.68   &45.43 &37.82 &33.55 \\
  OURS           &\pmb{94.64}   &\pmb{93.44} &\pmb{90.61} &\pmb{81.93}  &\pmb{88.24}   &\pmb{85.64} &\pmb{80.73} &\pmb{72.60}   &\pmb{88.79}   &\pmb{86.28} &\pmb{80.75} &\pmb{68.82}  \\                                                                             \hline
	\end{tabular}
\end{table*}
\begin{figure*}
	\centering
	\subfigure[Epoch 0 (NMI:63.78)]{
		\includegraphics[width=0.23\linewidth]{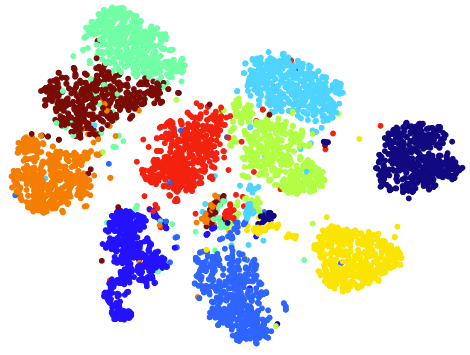}
	}
	\subfigure[Epoch 5 (NMI:91.15)]{
		\includegraphics[width=0.23\linewidth]{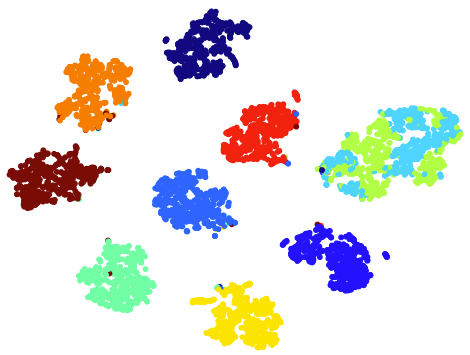}
	}
	\subfigure[Epoch 10 (NMI:97.47)]{
		\includegraphics[width=0.23\linewidth]{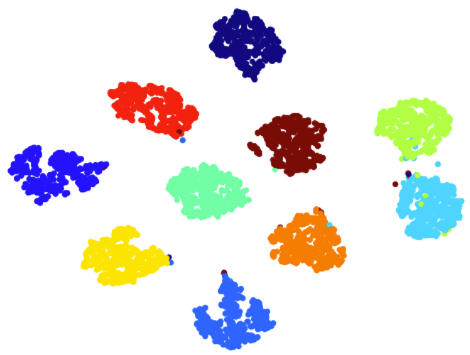}
	}
	\subfigure[Epoch 15 (NMI:98.23)]{
		\includegraphics[width=0.23\linewidth]{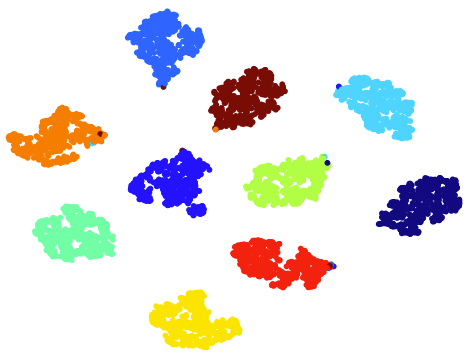}
	}
	\hspace{0.1 cm}
	\caption{t-sne visualization and NMI(\%) on the MNIST-USPS dataset for the contrastive learning process.}
	\label{fig:short1}
\end{figure*}

\begin{figure*}
	\centering
	\subfigure[Epoch 5 (NMI:82.14)]{
		\includegraphics[width=0.23\linewidth]{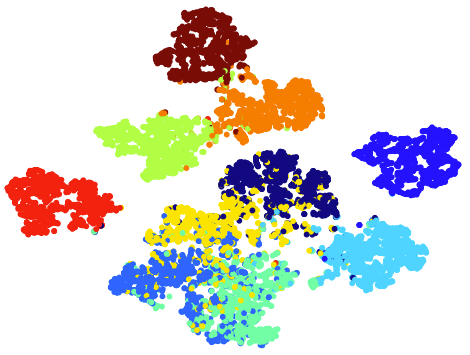}
	}
	\subfigure[Epoch 20 (NMI:90.91)]{
		\includegraphics[width=0.23\linewidth]{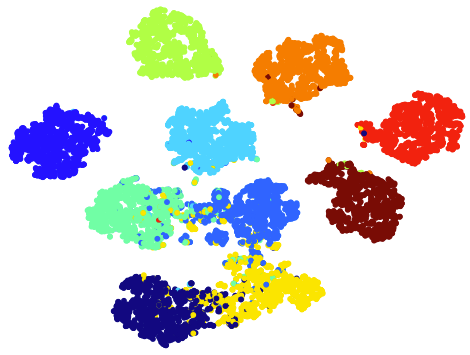}
	}
	\subfigure[Epoch 50 (NMI:92.57)]{
		\includegraphics[width=0.23\linewidth]{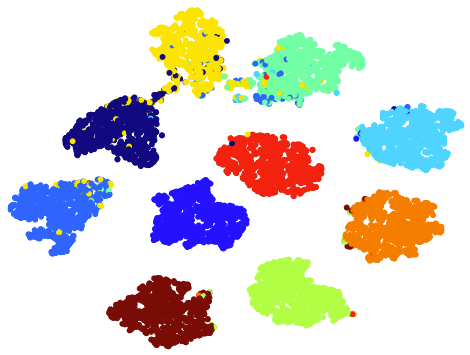}
	}
	\subfigure[Epoch 100 (NMI:92.86)]{
		\includegraphics[width=0.23\linewidth]{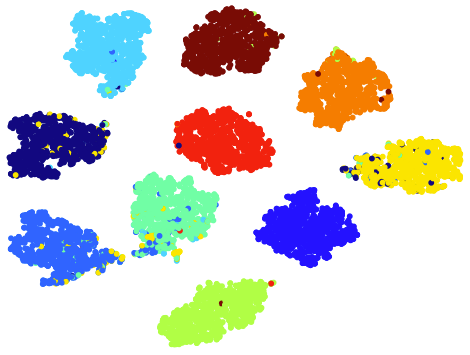}
	}
	\hspace{0.1 cm}
	\caption{t-sne visualization and NMI(\%) on the Fashion dataset for the contrastive learning process.}
	\label{fig:short2}
\end{figure*}

\subsection{Ablation Studies}
In this section, we will analyze the ablation studies of the proposed method on the MNIST-USPS dataset. In the evaluation, the missing rate is fixed at 0.5.

\begin{table}
	\caption{Ablation study of the loss part on the MNIST-USPS.}
	\label{tab:cluster7}\centering
\renewcommand\tabcolsep{4.2pt}
	\begin{tabular}{c|c|c|c|c|c}	
		\hline
$\mathcal{L}_{\mathrm{R}}$   & $\mathcal{L}_{\mathrm{C}}$  & $\mathcal{L}_{\mathrm{Z}}$ & ACC & NMI & ARI  \\ \hline
		  &   &  $\checkmark$   & 48.00 & 43.40     & 27.49 \\
		$\checkmark$   &  &    &47.14 & 44.83      & 32.42\\
		  & $\checkmark$  &     & 53.06 & 62.04       & 41.07\\
		$\checkmark$   &  & $\checkmark$    & 78.87 & 72.77      & 66.94 \\
		& $\checkmark$   & $\checkmark$     & 60.34 & 64.19      & 38.38 \\
		$\checkmark$  & $\checkmark$  &     & 83.22 & 80.58     & 74.21 \\
		$\checkmark$  & $\checkmark$ & $\checkmark$    & 97.48 & 93.54     & 94.49 \\\hline
	
	\end{tabular}
\end{table}
\begin{table}
	\centering
	\caption{Ablation study of the contrastive learning part on the MNIST-USPS}
	\label{tab:cluster8}\centering
\renewcommand\tabcolsep{4.2pt}
	\begin{tabular}{c|c|c|c|c}	
		\hline
$\mathbf{X}-\mathbf{Z}$   & $\mathbf{X}-\mathbf{Z}^*$  & ACC & NMI & ARI  \\ \hline
 $\checkmark$ &  & 81.40 & 72.48     & 65.94\\
$\checkmark$ & $\checkmark$  &89.72 & 81.07     & 79.05\\
	 & $\checkmark$  & 97.48 & 93.54    & 94.49\\\hline
	
	\end{tabular}
\end{table}

$\textbf{Loss function component analysis.}$ To further verify the importance of each module of our method in the overall algorithm framework, we evaluate each module of the algorithm through the following ablation studies.  In detail, as shown in Table \ref{tab:cluster7}, the following seven experiments are designed to isolate the effect of the contrastive loss $\mathcal{L}_{\mathrm{C}}$, the reconstruction loss $\mathcal{L}_{\mathrm{Z}}$, and the prediction loss $\mathcal{L}_{\mathrm{R}}$. It can be observed from the table that the loss of each part is very important for the model, no matter which part of the loss is missing will greatly affect the performance of the algorithm.

$\textbf{Contrastive learning on different representations.}$ To further illustrate the effectiveness of our direct contrastive learning model, we target and perform contrastive learning on different latent representations. As shown in the Table \ref{tab:cluster8},  we can find that the model performance is poor if both the reconstruction loss and the contrastive loss are applied to $\mathbf{Z}$. The main reason for the above situation is that the dataset contains a large amount of private information, and this method cannot handle the conflict between private information and common semantics, thus leading to performance degradation. And the performance advantage of our structure can be clearly observed, which demonstrates the effectiveness of our method on dealing with incomplete multi-view clustering problems.

\subsection{Visually Verify on Data Recovery}
In this section, we verify whether our method has the ability to restore the view through visual experiments. Experiments are carried out on the Noisy MNIST and Fashion datasets. By visualizing the recovered views, we can observe the effectiveness of the algorithm more intuitively. In the experiments, the missing rate $\eta$ is fixed at 0.5.

\begin{figure}
	\centering
	\includegraphics[width=0.9\linewidth]{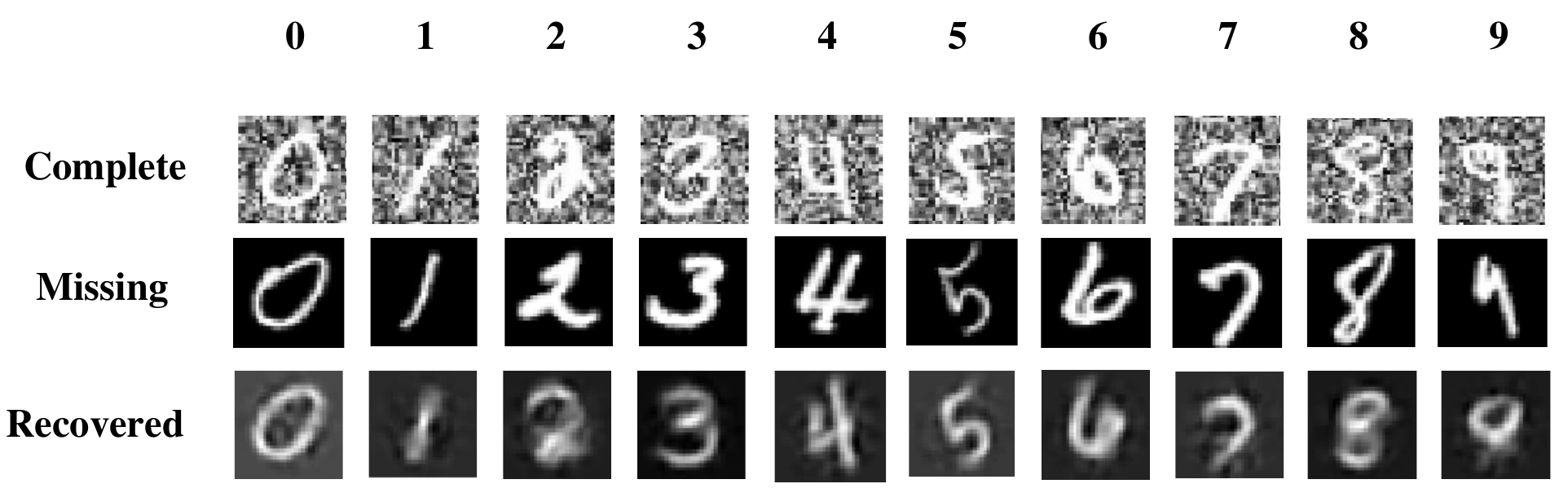}

	\includegraphics[width=0.9\linewidth]{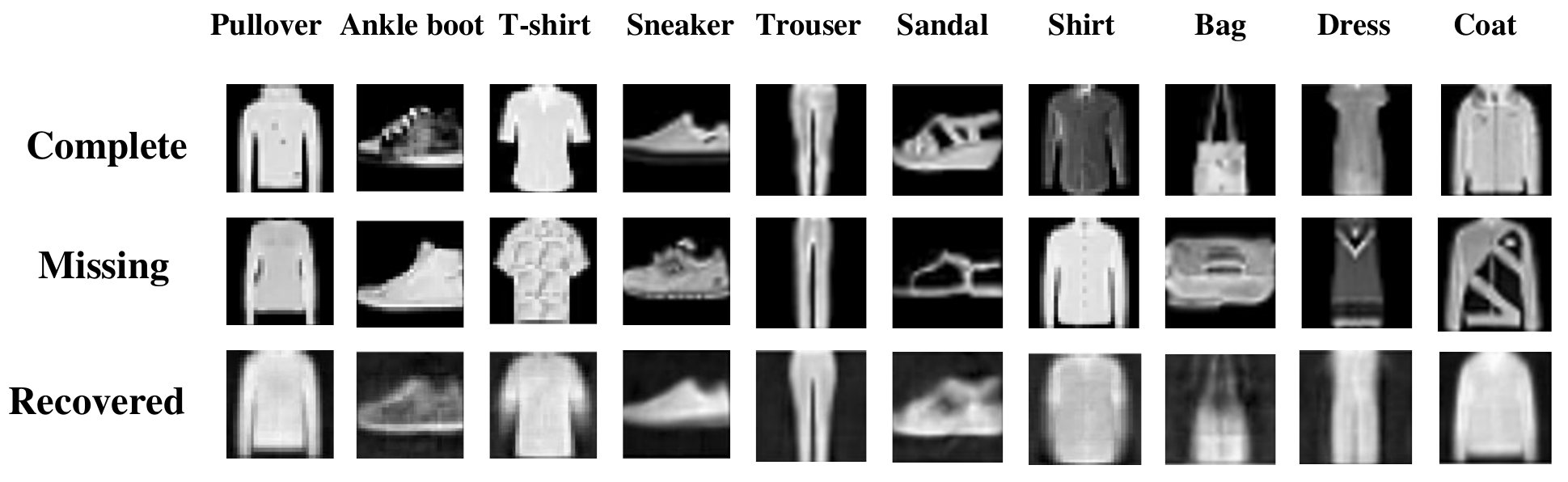}
	\caption{Data recovery on  Noisy MNIST and Fashion. Line 1 is the full view, Line 2 is the missing view, and Line 3 is the result of restoring from the full view.}
	\label{fig:short3}
\end{figure}

Figure \ref{fig:short3} demonstrates the data recovery ability of our method on the dataset Noisy MNIST and Fashion, and our method has the ability to infer missing view representations compared to existing state-of-the-art methods. As shown in the figure, we can observe that in the first three rows of data, the recovered images are very similar to the complete images in the first row, but the background of our newly acquired data is consistent with the lost view, so it can be concluded that our method can recover the important information of the missing view through the information of other views, and can discard the blurred noise in the original view.

\subsection{T-sne Visualizations}
In addition to the above visualizations, we also show t-sne visualizations of common representations learned by the model. As shown in the Figures \ref{fig:short1} and \ref{fig:short2}, as the epoch increases, the learned representation becomes more compact and discriminative.

\subsection{Parameter and Convergence Analysis}
\begin{figure}
	\centering
	\subfigure[Loss vs. performance on BDGP]{
		\includegraphics[width=0.45\linewidth]{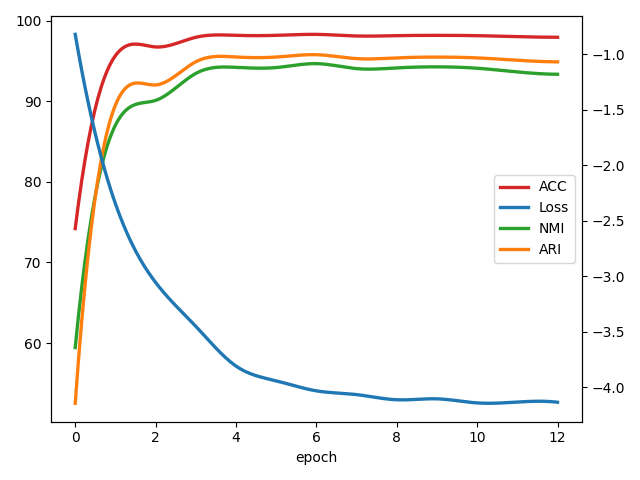}
	}
	\subfigure[Loss vs. performance on MNIST-USPS]{
		\includegraphics[width=0.45\linewidth]{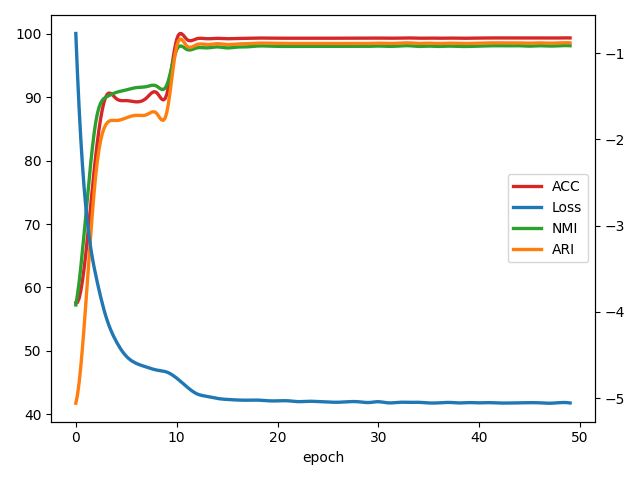}
	}
	\hspace{0.1 cm}
  \caption{Figure (a) is the result of the BDGP dataset , and Figure (b) is the result of the MNIST-USPS dataset.}
  \label{fig:short4}
\end{figure}

In this section, we investigate the convergence of our algorithm by reporting the loss value and the corresponding cluster performance over time. As shown in Figure \ref{fig:short4}, the x-axis represents training epochs, and the left and right y-axes represent clustering performance and corresponding loss values, respectively. It can be observed from the figure that the clustering effect of the model increases as the loss value decreases after several epochs, indicating that the proposed method has good convergence performance.

\begin{figure}
	\centering
	\subfigure[$\lambda_1$ vs. $\lambda_2$ (ACC)]{
		\includegraphics[width=0.45\linewidth]{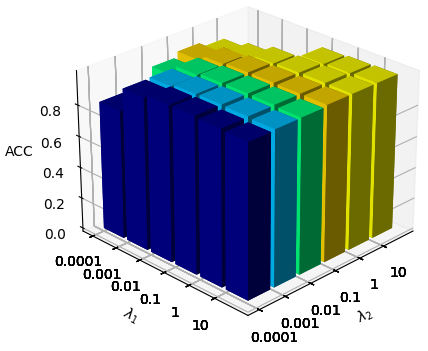}
	}
	\subfigure[$\lambda_1$ vs. $\lambda_2$ (NMI)]{
		\includegraphics[width=0.45\linewidth]{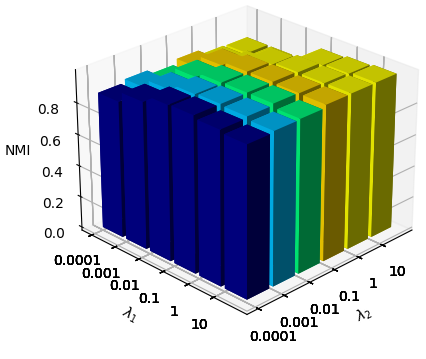}
	}
	\hspace{0.1 cm}
  \caption{Parameter analysis on MNIST-USPS.}
  \label{fig:short5}
\end{figure}

In the parameter sensitivity analysis, we studied the loss components in two hyper-parameters balance loss components, i.e., $\mathcal{L}_{\mathbf{Z}}+\lambda_1 \mathcal{L}_{\mathbf{C}}+\lambda_2 \mathcal{L}_{\mathbf{R}}$. Figure \ref{fig:short5} shows the average value of ACC and NMI for 5 independents runs, as shown in the figure, our method is robust to the choice of $\lambda_2$. In addition, a good choice of $\lambda_1$ will significantly improve the performance of our model.

\section{Conclusion}
\label{sec:conclusion}
To solve the dimensional collapse problem in incomplete multi-view clustering without relying on additional projection heads, this paper proposes a novel framework for contrastive learning of incomplete multi-view clustering. For each view, the framework performs reconstruction learning and consistency learning on the latent features and their sub-vectors respectively, which makes our model simpler to avoid dimensional collapse and guarantees that latent features can obtain more useful information. And recover the view by minimizing the conditional entropy while discarding inconsistent information. Since our method does not achieve consistency learning and reconstruction learning on the same feature, the conflict between view-private information inconsistency and common semantic consistency is well reduced. Extensive experiments on 5 public datasets demonstrate that our method achieves state-of-the-art performance.

\ifCLASSOPTIONcompsoc
  \section*{Acknowledgments}
\else
  \section*{Acknowledgment}
\fi

This work was supported in part by  the Fundamental
Research Funds for the Central Universities (No.31920220130, 31920220019), the Introduction of Talent Research Project of Northwest Minzu University (No.xbmuyjrc201904), the National Natural Science Foundation of China (No.61866033), the Leading Talent of National Ethnic Affairs Commission (NEAC), the Young Talent of NEAC, and the Innovative Research Team of NEAC (2018) 98.

\ifCLASSOPTIONcaptionsoff
  \newpage
\fi

{\small
\bibliographystyle{IEEEtran}
\bibliography{FM_ref}
	\vspace{-40pt}
\begin{IEEEbiography}[{\includegraphics[width=1in,height=1.25in,clip,keepaspectratio]{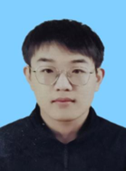}}]{Kaiwu Zhang}
  received the B.S. degree in Software Engineering from the College of the Dalian Maritime University, Dalian, China, in 2020. He is currently pursuing the M.S. degree with the College of China National Institute of Information Technology, Northwest Minzu University, Lanzhou, China. His current research interests include contrastive learning, and multi-view clustering.
\end{IEEEbiography}
\vspace{-40pt}
\begin{IEEEbiography}[{\includegraphics[width=1in,height=1.25in,clip,keepaspectratio]{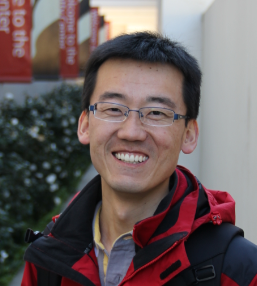}}]{Shiqiang Du}
	received the Ph.D. degree in radio physics from School of Information Science and Engineering at Lanzhou University, Lanzhou, China, in 2017. He was a visiting scholar with the Department of Mathematics and Statistics, Utah State University, Logan, USA, in 2013. From 2019 to 2020, he was a visiting scholar with the Department of Engineer, University of Modena and Reggio Emilia, Modena, Italy. He is currently the Professor with the College of Mathematics and Computer Science, Northwest Minzu University, Lanzhou, China. His current research interests include machine learning, image processing, and pattern recognition.
\end{IEEEbiography}
\vspace{-40pt}
\begin{IEEEbiography}[{\includegraphics[width=1in,height=1.25in,clip,keepaspectratio]{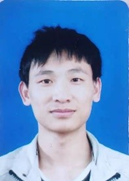}}]{Baokai Liu}
	 received the B.S. degree in communication engineering from the Shanghai Second Polytechnic University, Shanghai, China, in 2016. He is currently pursuing the M.S. degree with the College of China National Institute of Information Technology, Northwest Minzu University, Lanzhou, China. His current research interests include image processing, and deep learning.
\end{IEEEbiography}
\vspace{-40pt}
\begin{IEEEbiography}[{\includegraphics[width=1in,height=1.25in,clip,keepaspectratio]{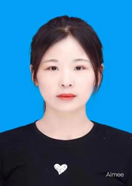}}]{Shengxia Gao}
	 received the B.S. degree in software engineering from the College of Mathematics and Computer Scinence, Northwest Minzu University, Lanzhou, China, in 2020. She is currently pursuing the M.S. degree with the College of China National Institute of Information Technology, Northwest Minzu University, Lanzhou, China. Her current research interests include contrastive learning.
\end{IEEEbiography}

\end{document}